\documentclass[10pt,twocolumn,letterpaper]{article}

\usepackage{3dv}
\usepackage{times}
\usepackage{epsfig}
\usepackage{graphicx}
\usepackage{amsmath}
\usepackage{amssymb}
\usepackage[english]{babel}

\usepackage[utf8]{inputenc} 
\usepackage[T1]{fontenc}    
\usepackage{hyperref}       
\usepackage{url}            
\usepackage{booktabs}       
\usepackage{amsfonts}       
\usepackage{nicefrac}       
\usepackage{microtype}      

\usepackage{caption}
\usepackage{floatrow}
\usepackage{amsmath}
\usepackage{mathtools}
\usepackage{textcomp}
\usepackage{wrapfig}
\usepackage{graphicx}
\graphicspath{{figures/}}
\usepackage[ruled]{algorithm2e} 

\SetAlFnt{\small}
\SetAlCapFnt{\small}
\SetAlCapNameFnt{\small}
\SetAlCapHSkip{0pt}
\IncMargin{-\parindent}

\usepackage{algpseudocode}
\usepackage{multirow}
\usepackage{enumitem}
\usepackage{color}

\threedvfinalcopy 

\def\threedvPaperID{155} 

\ifthreedvfinal\fi
\begin{document}

\title{DV-ConvNet: Fully Convolutional Deep Learning on Point Clouds \
 with Dynamic Voxelization and 3D
Group Convolution}

\author{
  Zhaoyu Su \quad Pin Siang Tan \quad Junkang Chow \quad Jimmy Wu \quad Yehur Cheong \quad Yu-Hsing Wang\\
  DESR Lab, Hong Kong University of Science and Technology 
}


\maketitle

\begin{abstract}
3D point cloud interpretation is a challenging task due to the randomness and sparsity of the component points. Many of the recently proposed methods like PointNet and PointCNN have been focusing on learning shape descriptions from point coordinates as point-wise input features, which usually involves complicated network architectures. In this work, we draw attention back to the standard 3D convolutions towards an efficient 3D point cloud interpretation. Instead of converting the entire point cloud into voxel representations like the other volumetric methods, we voxelize the sub-portions of the point cloud only at necessary locations within each convolution layer on-the-fly, using our dynamic voxelization operation with self-adaptive voxelization resolution. In addition, we incorporate 3D group convolution into our dense convolution kernel implementation to further exploit the rotation invariant features of point cloud. Benefiting from its simple fully-convolutional architecture, our network is able to run and converge at a considerably fast speed, while yields on-par or even better performance compared with the state-of-the-art methods on several benchmark datasets.

\end{abstract}

\section{Introduction}

Point cloud interpretation with deep learning is increasingly important with the growth of the autonomous driving industry. Convolutional Neural Network (CNN) \cite{LeCun2015,lecun1998gradient}, a staple deep learning technique, is known to exploit the strong spatially local correlation in objects. However, standard CNNs are only amenable to data defined over regular grids, and pioneering attempts \cite{maturana2015voxnet} in retrofitting CNN methods first utilized volumetric representation, i.e., discretized point cloud into 3D occupancy grid (voxel), then applied 3D convolutions. Initially, volumetric representation was considered highly advantageous as the method naturally preserves the neighborhood structure of 3D point clouds, which allowed direct application of standard 3D convolutions. However, point clouds are naturally sparse, and occupied grids \textemdash those with non-zero feature values \textemdash only account for a small percentage of the volume. To apply dense convolution neural networks on the spatially-sparse point cloud data is inefficient. 

\begin{figure}[t]
    \includegraphics[width=1.02\columnwidth]{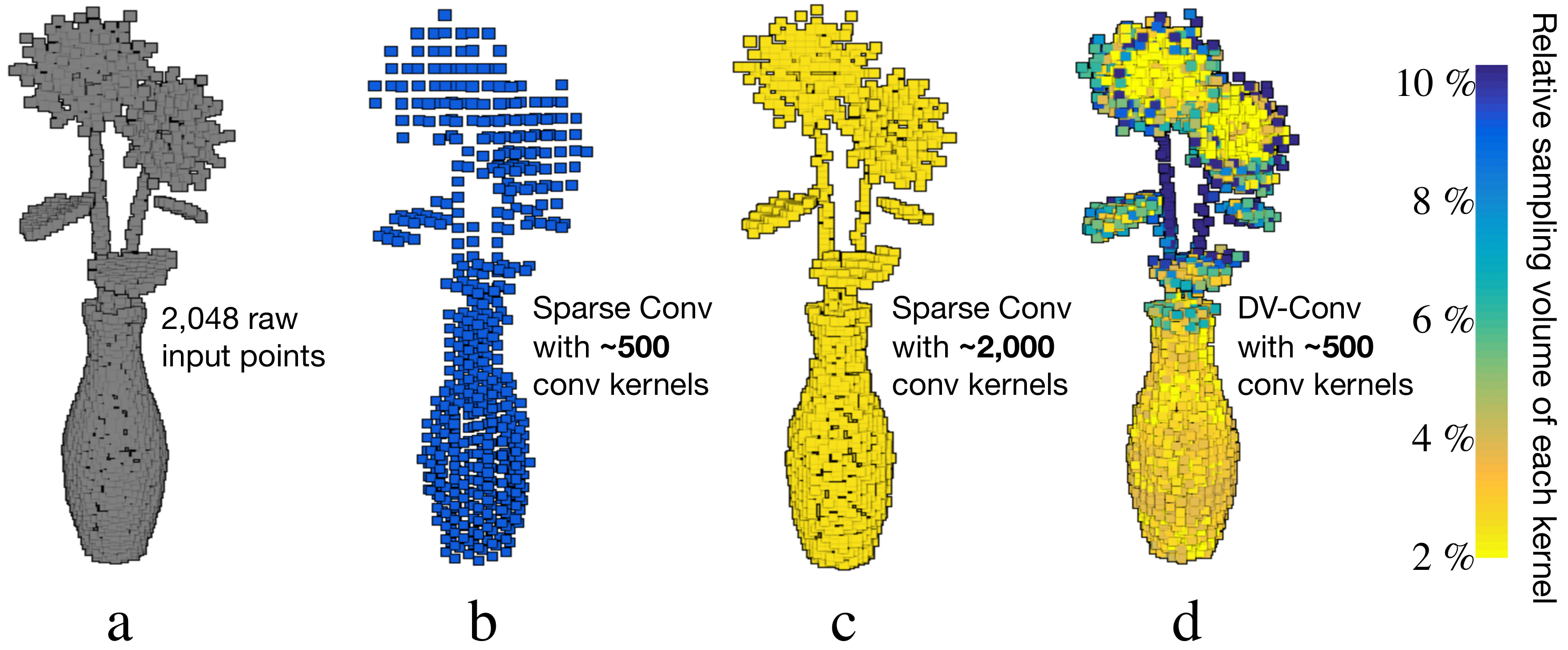}
    \centering
    {\caption{Visualization comparison between global voxelization (b, c) and our dynamic voxelization (d) with $3\times3\times3$ kernel size, given the input point cloud (a). Sparse convolution has already been taken into consideration, and the color map indicates the relative sampling volume of each convolution kernel, where brighter means finer.}\label{voxels}}
\end{figure}

Sparse convolutional networks-based methods were then introduced \cite{Graham2018,graham2017submanifold,choy20194d}, in attempts to resolve the problem. These methods significantly reduce memory and computational costs by restricting the convolution output to be only related to occupied voxels. However, the operation of quantization does not only handle point cloud data, since it naturally consists of non-uniform density in the 3D Euclidean space.
As such, the global voxelization process is a trade-off between the desired resolution and computation cost, and more often than not, the choice of faster computation will result in quantization artifacts, as illustrated in Fig. 1b. One can always increase the fineness of the grid; however, such an option increases the computation cost and decreases the sampling volume, i.e., the volume which is processed by a CNN kernel, as demonstrated in Fig.1c. Attempts of applying non-uniform grids have been reported, with examples such as the Kd-tree method \cite{Klokov2017} and Octree based method \cite{Wang,Riegler,Wang-2018-AOCNN,OctreeLiu2018} which use hierarchical structures that recursively partition a given volume to enable fine grids on parts that consist of denser point clouds. Unfortunately, these methods created data structures that are difficult to be implemented efficiently on a GPU. As such, in practice, the selection of appropriate grid resolution is non-trivial and may lead to very different outcomes, which is clearly undesirable.

Some researchs suggest adopting a non-volumetric approach to avoid the quantization artifact as a result of voxelization. 
In such methods, pointwise MLPs networks\cite{Qi2017a,Qi2017,Li2018} directly use the [x, y, z] coordinates of points as input features, learning point-wise features via shared multi-layer perceptrons (MLPs). Since these features are learned spatially independently from points coordinates, the contextual neighborhood structure of points cannot be preserved in the pointwise MLPs networks.
Further, continuous convolution kernels \cite{Xu2018,Wu}, which define convolutional kernels on continuous space, where the weights for neighboring points are related to the spatial distribution with respect to the center point, were proposed to facilitate the discernment of the context of neighboring points \cite{guo2019deep}.
However, continuous space computation causes inherent inefficiency on the GPU architecture due its non-discrete nature.
The inadequacy of the non-volumetric approach in balancing the geometry context comprehension and computational efficiency prompts us to consider a pragmatically comprehensive and computationally economic approach.

In this work, we would like to propose a method which simultaneously:
1) Has no need for a global voxelization. There should be no tradeoff between the desired resolution and computation cost; the resolution and computation efficiency should be balanced;
2) Preserves the neighborhood structure and context of 3D point clouds with spatially stable volumetric representation, and
3) Allows direct application of standard 3D convolutions.

We design DV-ConvNet, a volumetric convolution based method that employs dense 3D convolution in a point-wise fashion.
The core of DV-ConvNet is DV-Conv operator, which comprises two basic operations:

\begin{itemize}
\item \textbf{Dynamic voxel sampling}.
As shown in Fig. \ref{voxels}d, our method is able to construct convolutional kernels at location where there is point cloud, enabling us to preserve similar amount of details compared to Fig. \ref{voxels}c, while maintaining one quarter of the effective convolutional kernels (i.e., the same computing cost) as the operational outcome illustrated in Fig. \ref{voxels}b.
With self-adaptive sampling volume of each convolution kernel, our method, as shown in Fig.1d, is able to increase the sampling volume on sparsely populated areas, e.g., the stem, to save computation cost and decrease the receptive field on the floral disk and the vase so as to focus more on the fine geometry details.
By nature it is a differentiable operation converting regional sub-portions of point cloud into voxels on-the-fly during the model forward propagation to past information across layers to preserve the neighborhood structure of 3D point clouds and increase the receptive field. 

\item \textbf{3D Group convolution}, which is allowed by our dense 3D convolution kernel, extracting rotation invariant features well-suited for learning geometry in point clouds. 
\end{itemize}

We demonstrate DV-ConvNet in state-of-the-art and favorable performance for 3D shape classification and 3D point cloud semantic segmentation tasks, while maintaining high computational efficiency and implementation simplicity.

\section{Related Works}
\subsection{Convolution on Regular Grids}
CNN has become the go-to method for 2D image processing, and to extend CNN’s success to 3D point clouds, most of the CNN based methods simply convert the entire point cloud into voxel representations \cite{maturana2015voxnet,wang2015voting,qi2016volumetric,wu20153d}. However, the high sparsity of point clouds have limited both the resolution of the converted voxels and the depth of the CNN model, usually leading to a large computation waste on the void voxels. 

Therefore, sparse convolution \cite{Graham2018,graham2017submanifold,choy20194d} and sparse kernel \cite{Li2016} based methods are proposed by convolving only on the occupied voxels. On the other hand, variants of convolution which operate directly on more advanced data structures like the Oc-Tree \cite{Wang,Riegler,OctreeLiu2018} or Kd-Tree \cite{Klokov2017} are proposed, which are more effective for 3D data representation. VV-Net \cite{meng2019vv} is proposed, which employs a variational autoencoder (VAE) to learn a better representation of sparsity. In contrast, our method does not rely on global voxelization preprocessing. 

\subsection{Deep Learning with Non-volumetric Methods}
PointNet is the first neural network model that directly takes the raw points of a point cloud as input. It learns the point-order-invariant geometry feature representation from the point coordinates, using shared MLP with global max-pooling operation. To address the issue of local feature neglect, PointNet++ \cite{Qi2017} and SO-Net \cite{Li2018a} are proposed by using PointNet in a hierarchical fashion. Furthermore, to overcome the information loss issue induced by the global max-pooling, PointCNN \cite{Li2018} is proposed to achieve order equivariance through a learnable perturbation transformation matrix; similar ideas are also implemented in other related works \cite{Landrieu2017,dgcnn}. 

There are also methods extending the notion of convolution from regular grids to point sets in $\mathbb{R}^n$ by using MLPs \cite{Wu} or learnable Taylor polynomials \cite{Xu2018} to approximate a continuous convolution kernel, however, they usually involve massive non-linear computations, and therefore are intractable in practice and have limited applications. While in our method, we preserve the neighborhood structure and context of 3D point clouds with spatially stable volumetric representation.

\subsection{Group Convolution}
Transformation equivariance is important for 3D point cloud interpretation, and in previous works, this property was studied based on harmonic networks \cite{worrall2017harmonic,thomas2018tensor} or MLPs operated in Fourier space \cite{kondor2018n}, where the geometry of point cloud is still not explicitly modeled. 

On the other hand, in a traditional convolution, the kernel sliding operation with parameter sharing inherently ensures the equivariance to translations: shifting the input image and then feeding it to the convolution is equivalent to feeding the same input image to the convolution and shifting the output feature map afterwards \cite{Cohen2016}. 

Group convolution extends the sharing of kernel parameters to a larger extent by employing a symmetrical group of convolution kernels, which are equivariant to a wider range of geometric transformations such as rotation and reflection \cite{Cohen2016,Veeling2018}. Thus, a CNN equipped with group convolution possesses stronger transformation equivarant ability, and therefore is better at leveraging the image symmetry towards robust feature learning, without increasing the model size or computation overhead. 

CubeNet \cite{worrall2018cubenet} is the first work th extend group convolution from 2D images to 3D voxels; VV-Net \cite{meng2019vv} also employs group convolution, however, it is only applied at the latent feature space for one single convolution layer. In this work, we apply 3D group convolution in a end-to-end fashion based on $p4$ and $p4m$  symmetry groups, which are extended from their previously defined domain $\mathbb{Z}^2$ to  $\mathbb{Z}^3$. Working together with the proposed dynamic voxelization method, our group convolution based method achieves state-of-the-art performance. More details about group convolution are given in section \ref{3DVroup}, including its mathematical definition and implementation.

\begin{figure*}[h]
  \centering
  \includegraphics[width=0.9\textwidth]{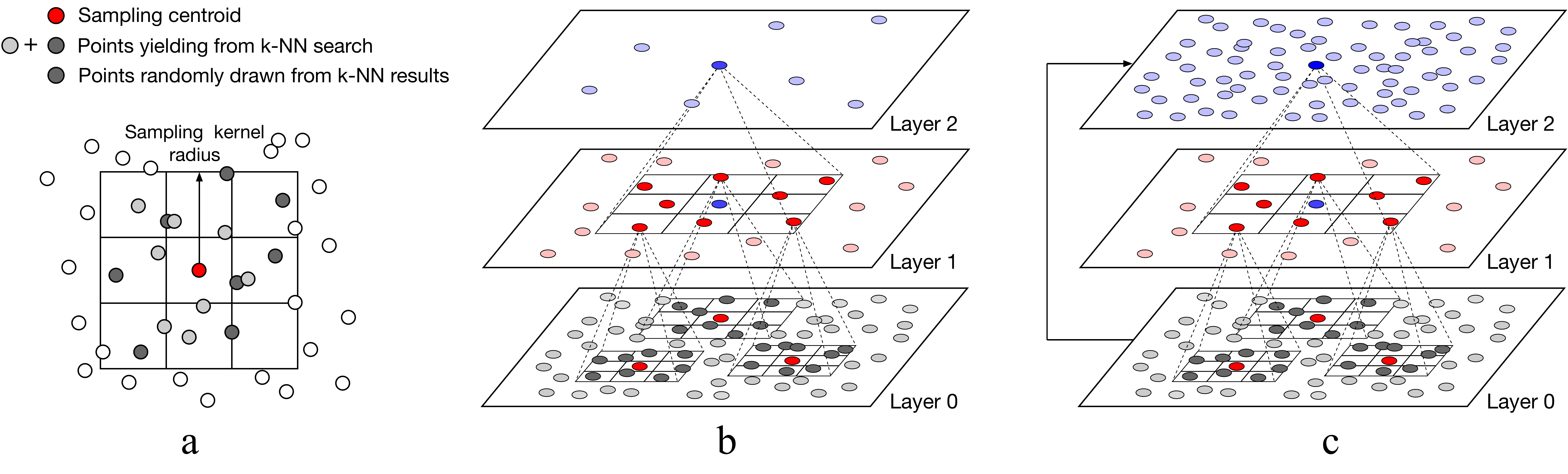}
  \caption{(a) Voxel sampling operation demonstrated in 2D case using $K$ = 8, $D$ = 2 and $S$ = 3. Here we choose $K$ = 8 for a better visualization, the value of $K$ used in practice is very large and there will be less ambiguity; (b) a 3-level hierarchy built upon G-Conv layers, notice that the sampling radius vary with the local point density; (c) a "Conv-Deconv" pair used in segmentation, the point coordinates in the convolution layer (Layer 0) are used as sampling centroids in its corresponding deconvolution pair (Layer 2).}
  \label{knn}
\end{figure*}

\section{DV-ConvNet}
We propose a new operator named DV-Conv, which aims at encoding the local geometry features of a point set into a 1D point-wise feature vector using 3D group convolution. Based on DV-Conv, we develop DV-ConvNet, a hierarchical deep neural network architecture for 3D point cloud interpretation. In this section, we first give a detailed introduction to the DV-Conv operator, including dynamic voxelization operation and 3D group convolution. Next, we briefly introduce two versions of DV-ConvNet for point cloud classification and segmentation.

\subsection{DV-Conv Operator}
\subsubsection{Dynamic Voxelization}
Given an input point cloud and a sampling centroid, we first draw $K\times D$ nearest neighbouring points (both dark grey and light grey points in Fig. \ref{knn}a) around the sampling centroid using the k-nearest-neighbours (k-NN) searching algorithm. Next, $K$ points (dark grey points in Fig. \ref{knn}a) are randomly picked among these $K\times D$ neighbours and are fitted into a cubic voxel kernel centering at the sampling centroid, while the kernel radius is determined such that the farthest neighbouring point can just be included. Thus, the receptive field of each convolution kernel can dynamically adapt to the local points density in each convolution layer.

Each sampling kernel consists of $S^3$ uniform regular grids ($S$ is known as kernel size in this context), while the representative feature of each grid is calculated based on the points falling into it. For a grid occupied by one single point, the feature of that point is treated as the feature representation; for a grid not occupied by any point, zero values are assigned for that grid following the practice in \cite{Graham2018}. If a grid is occupied by more than one point, a max-pooling strategy is applied, where the channel-wise maximum across all the interior points is used as the representative feature for that grid. This is analogous to the max-pooling method used in traditional CNNs, but ours occurs within each grid instead of over the entire kernel. Indices of the points that contribute to the grid features are recorded during the network forward propagation and they are used to guide the gradient backward propagation, therefore, our dynamic voxelization operation is fully differentiable. For each grid, only the first five points encountered during the forward propagation are taken into consideration, while the rest are neglected. Since the computation expense varies cubically along with kernel size $S$, to control the computation expense within a rational range, $S$ is fixed as 3 for all the DV-Conv layers. During the dynamic voxelization operation, $N$ sampling centroids are drawn from the input point cloud using the farthest point sampling method, thus, the dynamic voxelization operation converts a point cloud in $\mathbb{R}^{C_{in}}$ into $N\times S^3$ voxels in $\mathbb{R}^{C_{in}}$, where $C_{in}$ is the number of feature channels of the point cloud. 

We highlight the efficiency of the dynamic voxelization operation when compared with the global voxelization, which is commonly used in previous works \cite{maturana2015voxnet,wang2015voting,qi2016volumetric,wu20153d}. Though the invention of sparse convolution alleviates the computation inefficiency issue induced by the point cloud sparsity, how to choose the proper resolution for the global voxelization preprocessing is still a crucial yet challenging problem, since an improperly selected resolution may cause either severe information loss or massive computation overheads, as illustrated in Fig. \ref{voxels}b and c. In contrast, as shown in Fig. \ref{voxels}d, both the locations and receptive fields of the convolution kernels are more concentrated at the locations where the points are densely gathered, such as the floral discs and vase, and vice versa. Moreover, the introduction of dilation rate $D$ further adjusts the receptive fields to include richer features without additional computation expense, and in this way, the computation power is more efficiently allocated.

Specifically, \cite{Hua} shares the concept of point-wise convolution with our work, but the similarity ends there. Dynamic voxelization sampling is the key to solve the following architecture limitation of \cite{Hua} which severely hamper its real-world applications: \cite{Hua} is unable to organize into a hierarchical network architecture, because each layer in \cite{Hua} always shares the same receptive field. Unless dynamic voxelization sampling is introduce, the convolution layer in \cite{Hua} cannot adapt to the dynamic change of feature points, not to mention the encoder-decoder architecture for segmentation tasks. The performance on benchmark datasets in Table \ref{tab:ModelNet40} and \ref{tab:stat} further illustrates the inherent difference between \cite{Hua} and our work, despite their similar network architecture for the classification task.

\subsubsection{3D Group Convolution.}
\label{3dgroup}
Mathematically, the convolution between 3D feature voxels $\mathbf{F}$ and a 3D convolution kernel $\mathbf{W}$ can be expressed using Eq. \ref{eq:3DConv}:

\begin{equation}
\mathbf{W} \star \mathbf{F}=
\sum_{\mathbf{c}\in\mathbb{Z}^3} [t\mathbf{W}]_\mathbf{c}\mathbf{F}_\mathbf{c}
=\sum_{\mathbf{c}\in\mathbb{Z}^3} \mathbf{W}_{t^{-1}\mathbf{c}}\mathbf{F}_\mathbf{c}  \label{eq:3DConv}
\end{equation}

where $\star$ is the convolution operator, $\mathbf{c}$ is the coordinate of voxel in $\mathbf{F}$, $\mathbf{c}=[x,y,z]^T\in \mathbb{Z}^3$, $t$ stands for the translation for $\mathbf{W}$, and we omit the summation among feature channels for simplicity. Eq. \ref{eq:3DConv} can be interpreted as follows: shifting $\mathbf{W}$ to the convolution location $\mathbf{c}$ in $\mathbf{F}$ using translation $t$, is equivalent to finding the corresponding location of weight in $\mathbf{W}$ by using the inverse translation $t^{-1}$ on $\mathbf{c}$, that is $[t\mathbf{W}]_\mathbf{c}=\mathbf{W}_{t^{-1}\mathbf{c}}$. For any translation $s$ acts on $\mathbf{F}$, we have:

\begin{equation}\label{eq:ShiftEqui}
\begin{split}
\mathbf{W} \star [s\mathbf{F}]&=
\sum_{\mathbf{c}\in\mathbb{Z}^3} [t\mathbf{W}]_\mathbf{c}[s\mathbf{F}]_\mathbf{c}
=\sum_{\mathbf{c}\in\mathbb{Z}^3} \mathbf{W}_{t^{-1}\mathbf{c}}\mathbf{F}_{s^{-1}\mathbf{c}} \\
&=\sum_{\mathbf{c'}\in\mathbb{Z}^3} \mathbf{W}_{t^{-1}s\mathbf{c'}}\mathbf{F}_{\mathbf{c'}}
=[\mathbf{W} \star \mathbf{F}]_{s^{-1}}=s[\mathbf{W} \star \mathbf{F}]
\end{split}
\end{equation}

Note that we leverage the substitution: $s^{-1}\mathbf{c}=\mathbf{c}'$, and $\mathbf{c}'$ is still defined in $\mathbb{Z}^3$. Thus, a 3D convolution is equivariant to translation by its definition. 

Now we extend the summation domain from $\mathbf{c}\in \mathbb{Z}^3$ to $g \in G$ as illustrated in Eq. \ref{eq:3DGroupConv}, where $G$ is a group of transformations which are equivariant to a wider range of transformations like rotation and reflection, and $G$ is also defined in $\mathbb{Z}^3$. 

\begin{equation}
\mathbf{W} \star \mathbf{F}=
\sum_{g\in G} [t\mathbf{W}]_g\mathbf{F}_g
=\sum_{g\in G} \mathbf{W}_{t^{-1}g}\mathbf{F}_g \label{eq:3DGroupConv}    
\end{equation}

Similarly, the conclusion like Eq. \ref{eq:ShiftEqui} still holds, which means, for any $g' \in G$, $\mathbf{W} \star [g'\mathbf{F}]=g'[\mathbf{W} \star \mathbf{F}]$, in other words, the group convolution is equivariant to the transformation group $G$. In this work, we employ $p4$ and $p4m$ symmetry groups as our $G$s, and extend them to $\mathbb{Z}^3$ from their originally defined domain $\mathbb{Z}^2$.

\textbf{$\mathbf{p4}$ / $\mathbf{p4m}$ Group.} The $p4$ group comprises a group of rotation transformations by $0$, $\pi / 2$, $\pi$, and $3\pi / 2$, regarding any rotation center in a square grid, and it is easy to conclude that $p4$ is a symmetry  group regrading rotation by $\pi/2$. Expressing $p4$ symmetry group in a homogeneous coordinate system, gives:

\begin{equation}
G_{p4}=
\begin{bmatrix}
\cos(r\pi/2) & -\sin(r\pi/2) & x\\
\sin(r\pi/2) & \cos(r\pi/2) & y\\
0 & 0 & 1\\
\end{bmatrix}   \label{eq:P4} 
\end{equation}

where  $r\in\left\{0,1,2,3 \right\}$ and $(x,y)\in \mathbb{Z}^2$. To extend the $p4$ group to $\mathbb{Z}^3$, we generalize its working domain from a square grid to cubic space by adding an additional $z$-axis. Putting the rotation center at the centroid of the cube and keeping the rotation at the $x$-$y$ plane, Eq. \ref{eq:P4} can be rewritten as:

\begin{equation}
G_{p4}=
\begin{bmatrix}
\cos(r\pi/2) & -\sin(r\pi/2) & 0 & 0\\
\sin(r\pi/2) & \cos(r\pi/2) & 0 & 0\\
0 & 0 & 1 & 0\\
0 & 0 & 0 & 1\\
\end{bmatrix} \label{eq:3DP4}
\end{equation}

The $p4m$ group extends the $p4$ symmetry by involving additional horizontal mirroring (reflection) transformations:

\begin{equation}
G_{p4m}=
\begin{bmatrix}
(-1)^m \cos(r\pi/2) & -(-1)^m\sin(r\pi/2) & 0 & 0\\
\sin(r\pi/2) & \cos(r\pi/2) & 0 & 0\\
0 & 0 & 1 & 0\\
0 & 0 & 0 & 1\\
\end{bmatrix} \label{eq:3DP4M}
\end{equation}

where $m \in \left\{0,1\right\}$. When $m=0$, $p4m$ group reduces to $p4$. 

\begin{figure}[t]
    \includegraphics[width=0.9\columnwidth]{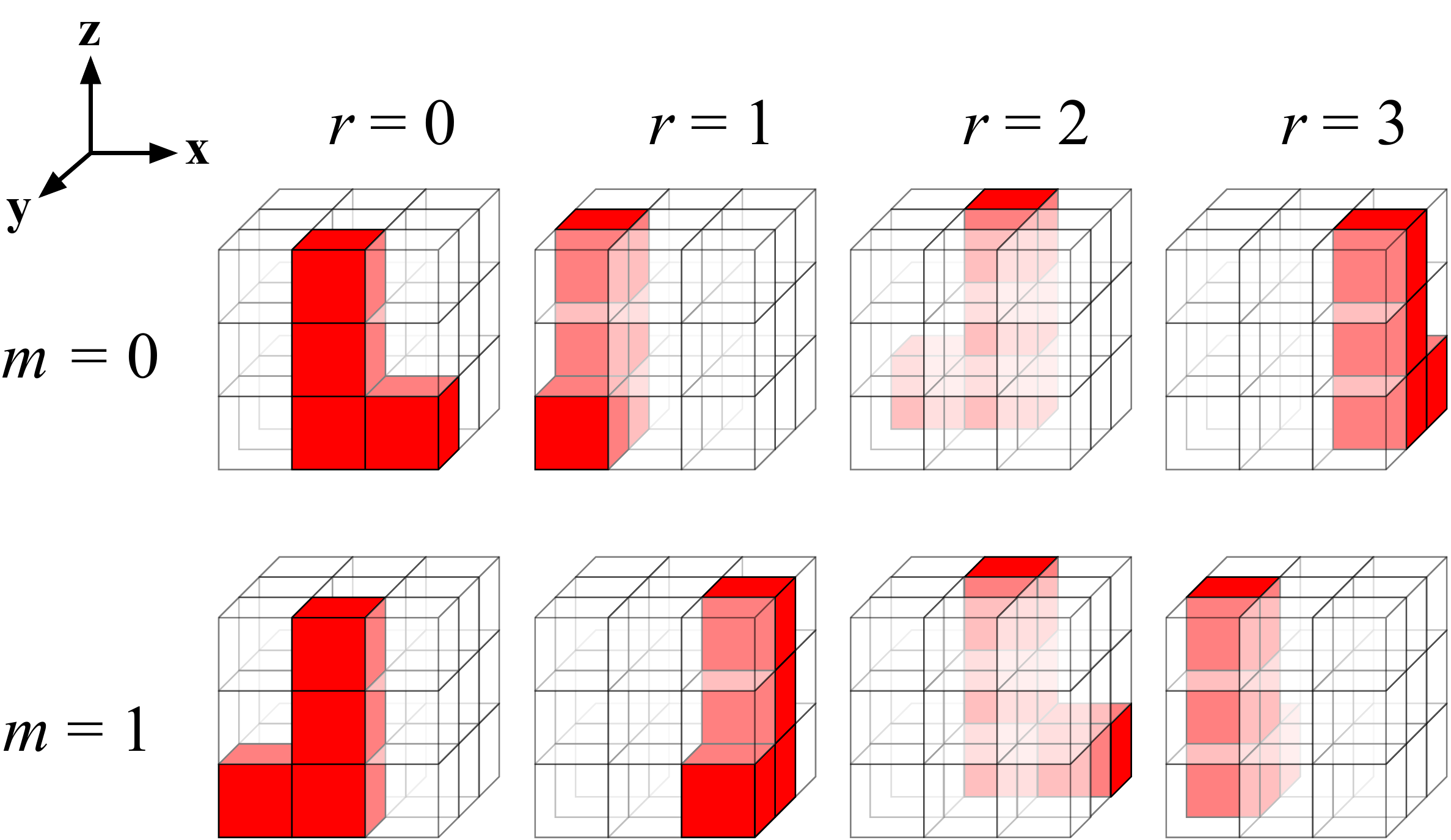}
    \centering
    {\caption{Visualization of 3D convolution kernels transformed via $p4$ group (the first row only) and $p4m$ group (both two rows) based on different transformation parameters $r$ and $m$.  }\label{kernels}}
\end{figure}

\begin{figure*}[h]
  \centering 
  \includegraphics[width=1.0\textwidth]{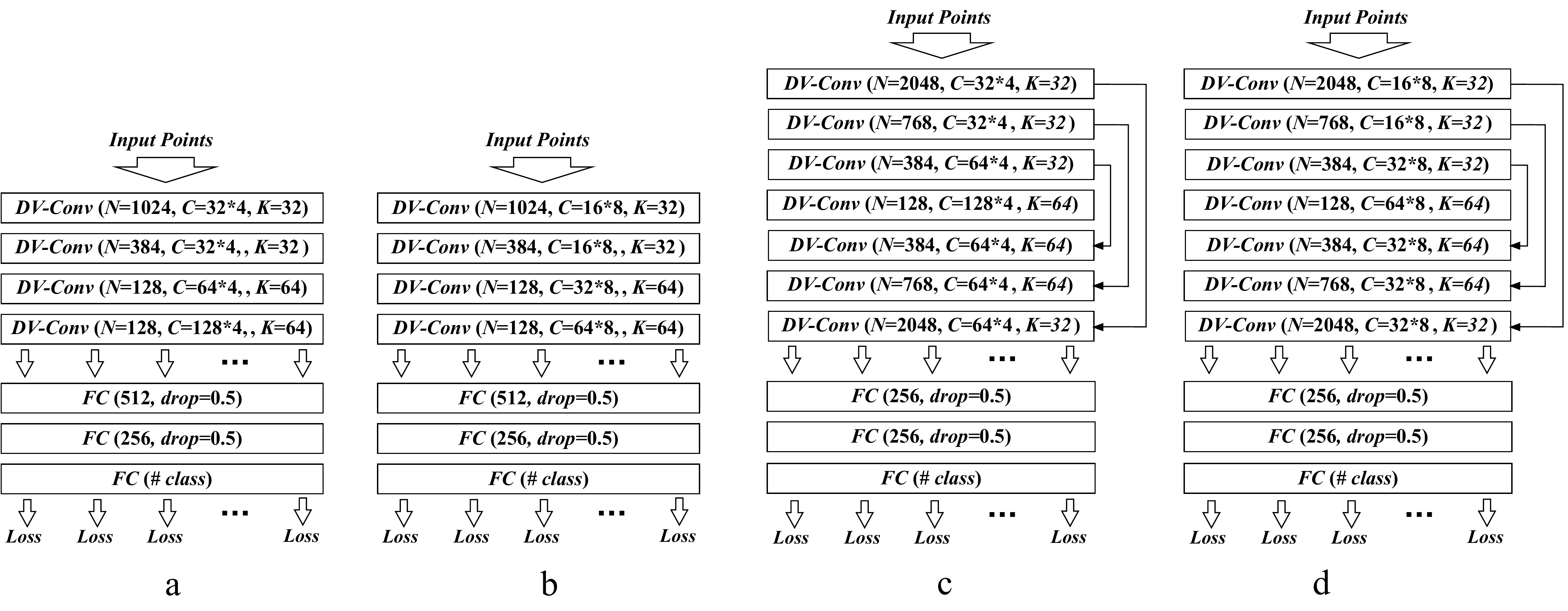}
  \caption{(a,b): $p4$ and $p4m$ based architectures for classification; (c,d): $p4$ and $p4m$ based architectures for segmentation. We note the number of channels as $C=*\times4$ or $C=*\times8$ to accommodate the kernel duplication in $p4$ and $p4m$ group convolution.}
  \label{architecture}
\end{figure*}

\textbf{Implementation.} Given the feature voxels $\mathbf{F} \in \mathbb{R}^{N\times S^3\times C_{in}}$ yielded from the dynamic voxelization operation, we define a convolution kernel $\mathbf{W}\in \mathbb{R}^{S^3\times C_{in}\times C_{out}}$ to convolve with $\mathbf{F}$, where $C_{in }$ and $C_{out}$ refer to the number of feature channels for input and output. According to the definition of group convolution, for each $g \in G$, we are able to transform the kernel $\mathbf{W}$ to $\mathbf{W}_g$ using $\mathbf{W}_g=g \mathbf{W}$. We use the notation $\mathbf{W}_{G} \in \mathbb{R}^{S^3\times C_{in}\times C_{out}\times n}$ to represent a concatenation of $\mathbf{W}_g$ transformed via all the $g \in G$, where $n$ is the total number of transformations $g$ in group $G$, $n=4$ for ${G_{p4}}$ and $n=8$ for ${G_{p4m}}$. A visualization of $\mathbf{W}_{G_{p4}}$ and $\mathbf{W}_{G_{p4m}}$ is given in the Fig.\ref{kernels} regarding different transformation parameters $r$ and $m$ in Eq. \ref{eq:3DP4M}.

\subsection{DV-ConvNet Architecture}



We propose two hierarchical network architectures for point cloud classification and segmentation based on a recursive application of the DV-Conv layers, and we name our network as DV-ConvNet. DV-ConvNet has a similar network design when compared to traditional CNNs for 2D images interpretation, they both consist of a hierarchy of convolutions followed by MLPs. Fig. \ref{knn}b presents a 3-level hierarchy of DV-Conv layers for classification, where the output from one DV-Conv layer serves as input for its successive DV-Conv layer. 

The DV-ConvNet architecture for segmentation follows the widely adopted encoder-decoder design of U-Net \cite{Ronneberger2015}. Fig. \ref{knn}c presents a “Conv-Deconv” pair, where the deconvolution part basically is still a DV-Conv layer, but in order to raise the number of returning points for higher output resolution, the coordinates of points in the corresponding encoder part are treated as sampling centroids for dynamic voxelization.


\section{Experiments}
We evaluate our DV-ConvNet on ModelNet40 dataset \cite{wu20153d} for point cloud classification, and evaluate on the ShapeNet-part \cite{shapenet2015} and S3DIS \cite{armeni_cvpr16} datasets for semantic segmentation. 
The detailed DV-ConvNet architectures of $p4$ and $p4m$ based models for classification and segmentation tasks are illustrated in the Fig.\ref{architecture}. Note that to keep a comparable computation expense, we halve the number of feature channels in $p4m$ based networks.

\subsection{Results on Benchmark Datasets}

\subsubsection{Classification}  We evaluate DV-ConvNet on ModelNet40 \cite{wu20153d} with a 9,843 / 2,468 training / testing split, and to make a fair comparison, we evaluate our model based on both “pre-aligned” and “unaligned” mode, following the practice in \cite{Li2018}. We use the dataset provided in \cite{Qi2017a}, and randomly select 1,024 points with their normals as our input point cloud. Anisotropic triaxial scaling in the range $[0.95, 1.05]$ is used for data augmentation in both evaluation modes. The test results and comparisons with other works are given in Table \ref{tab:ModelNet40}. It can be seen that DV-ConvNet equipped with $p4$ group achieves the best performance among all the other methods on both “pre-aligned” and “unaligned” evaluation modes. The reason why $p4m$ based network performs less well is due to the number of trainable parameters being only a half of the network equipped with $p4$ group. 

\begin{table}[h!]
\centering
  \scalebox{0.85}{
    \begin{tabular}{  l | c | c  }
             \cline{1-3}
             Method & Pre-aligned & Unaligned  \\
      \hline
      SO-Net~\cite{Li2018a} & 90.9 & -  \\
      Pointwise CNN~\cite{Hua} & 86.1 & -   \\
      Kd-Net~\cite{Klokov2017} & 90.6 & -   \\
            PCNN~\cite{AtzmonML18} & 92.3 & -   \\
            3DmFV-Net~\cite{ben20183dmfv} & 91.4 & -  \\
            A-CNN~\cite{komarichev2019acnn} & 92.6 & -   \\
            PointCNN~\cite{Li2018} & 92.5 & 92.2  \\
            PointNet~\cite{Qi2017a} & - & 89.2   \\
            PointNet++~\cite{Qi2017} & - & 90.7   \\
            SpiderCNN~\cite{Xu2018} & - & \textbf{92.4} \\
            Spec-GCN~\cite{Wang2018} & - & 91.5  \\
            DGCNN~\cite{dgcnn} & - & 92.2 \\
            \hline
            Ours ($p4$)~ &  \textbf{93.3} & \textbf{92.4}  \\
      Ours ($p4m$)~ & \textbf{93.2} & 92.3 \\
      \hline
    \end{tabular}
  }
    \caption{Classification result comparisons on ModelNet40 in overall accuracy (OA, \%) with "Pre-aligned" and "Unaligned" evaluation mode}
  \label{tab:ModelNet40}
\end{table}

\begin{table*}[h!]
  \centering
  \scalebox{0.71}{
    \begin{tabular}{  l | c | c | c | c | c | c | c | c | c | c | c | c | c | c | c | c | c | c  }
             \cline{1-19}
             Method & mIoU & mpIoU & air- & bag & cap & car & chair & ear- & guitar & knife & lamp & laptop & motor- & mug & pistol & rocket & skate- & table\\
             &  &  & plane &  &  &  &  & phone &  &  &  &  & bike &  &  &  & board & \\
      \hline
      PointNet~\cite{Qi2017a} & 80.4 & 83.7 & 83.4 & 78.7 & 82.5 & 74.9 & 89.6 & 73.0 & 91.5 & 85.9 & 80.8 & 95.3 & 65.2 & 93.0 & 81.2 & 57.9 & 72.8 & 80.6\\
      PointNet++~\cite{Qi2017} & 85.1 & 81.9 & 82.4 & 79.0 & 87.7 & 77.3  & 90.8  & 71.8 & 91.0 & 85.9  & 83.7 & 95.3 & 71.6 & 94.1 & 81.3 & 58.7 & 76.4 & 82.6 \\
      SSCN~\cite{Graham2018} & 86.0 & 83.3 & 84.1 & 83.0 & 84.0 & \textbf{80.8} & \textbf{91.4} & 78.2 & 91.6 & \textbf{89.1} & 85.0 & 95.8 & 73.7 & 95.2 & 84.0 & 58.5 & 76.0 & 82.7 \\
            SpiderCNN~\cite{Xu2018} & 85.3 & 81.7 & 83.5 & 81.0 & 87.2 & 77.5 & 90.7 & 76.8 & 91.1 & 87.3 & 83.3 & 95.8 & 70.2 & 93.5 & 82.7 & 59.7 & 75.8 & 82.8 \\
            RSNet~\cite{huang2018recurrent} & 84.9 & 81.4 & 82.7 & 86.4 & 84.1 & 78.2 & 90.4 & 69.3 & 91.4 & 87.0 & 83.5 & 95.4 & 66.0 & 92.6 & 81.8 & 56.1 & 75.8 & 82.2 \\
            SO-Net~\cite{Li2018a} & 84.9 & 81.0 & 82.8 & 77.8 & 88.0 & 77.3 & 90.6 & 73.5 & 90.7 & 83.9 & 82.8 & 94.8 & 69.1 & 94.2 & 80.9 & 53.1 & 72.9 & 83.0 \\
            VV-Net~\cite{meng2019vv} & 86.0 & 81.9 & 82.1 & 68.9 & 83.8 & 80.9 & 87.8 & 81.2 & 91.2 & 78.4 & 77.4 & 94.5 & 72.8 & \textbf{98.0} & \textbf{86.0} & 53.8 & \textbf{83.9} & \textbf{90.0} \\
            PointCNN~\cite{Li2018} & \textbf{86.1} & \textbf{84.6} & 84.1 & \textbf{86.5} & 86.0 & \textbf{80.8} & 90.6 & 79.7 & \textbf{92.3} & 88.4 & 85.3 & 96.1 & \textbf{77.2} & 95.3 & 84.2 & 64.2 & 80.0 & 83.0 \\
            A-CNN~\cite{komarichev2019acnn} & \textbf{86.1} & 84.0 & \textbf{84.2} & 84.0 & 88.0 & 79.6 & 91.3 & 75.2 & 91.6 & 87.1 & \textbf{85.5} & 95.4 & 75.3 & 94.9 & 82.5 & 67.8 & 77.5 & 83.3 \\
            
      \hline
      Ours ($p4$) & \textbf{86.1} & \textbf{84.6} & 82.6 & 86.2 & \textbf{89.7} & 80.0 & 89.4 & \textbf{82.6} & 92.0 & 87.9 & 78.7 & \textbf{96.6} & 74.3 & 94.9 & 83.6 & \textbf{69.0} & 78.7 & 87.8\\
      Ours ($p4m$) & 85.9 & 84.2 & 81.9 & 85.0 & 88.9 & 80.4 & 89.4 & 82.7 & 91.7 & 88.2 & 77.8 & 95.5 & 73.6 & 95.1 & 83.8 & \textbf{68.6} & 77.6 & 87.0\\
      
      \hline
    \end{tabular}
  }
  \caption{Evaluation results on ShapeNet-part in part-averaged IoU (mIoU, \%) , mean per-class pIoU (mpIoU, \%) and per-class pIoU (\%).}
  \label{tab:ShapeNet}
\end{table*}

\begin{table*}[h!]
  \centering
  \scalebox{0.71}{
        \begin{tabular}{  l | c | c | c | c | c | c | c | c | c | c | c | c | c | c | c | c }
             \cline{1-17}
             Method & OA & mAcc & mIoU & ceiling & floor & wall & beam & column & window & door & table & chair & sofa & bookcase & board & clutter \\
      \hline
            PointNet~\cite{Qi2017a} & - & 48.98 & 41.09 & 88.80 & 97.33 & 69.80 & 0.05 & 3.92 & 46.26 & 10.76 & 58.93 & 52.61 & 5.85 & 40.28 & 26.38 & 33.22 \\
            SPGraph~\cite{Landrieu2017} & 86.38 & 66.50 & 58.04 & 89.35 & 96.87 & 78.12 & 0.00 & \textbf{42.81} & 48.93 & 61.58 & 84.66 & 75.41 & 69.84 & 52.60 & 2.10 & 52.22 \\
            PointCNN~\cite{Li2018} & 85.91 & 63.86 & 57.26 & 92.31 & 98.24 & 79.41 & 0.00 & 17.60 & 22.77 & 62.09 & 74.39 & 80.59 & 31.67 & 66.67 & 62.05 & 56.74 \\
            Minkowski~\cite{choy20194d} & - & 71.71 & \textbf{65.35} & 91.75 & \textbf{98.71} & \textbf{86.19} & 0.00 & 34.06 & 48.90 & \textbf{62.44} & \textbf{81.57} & \textbf{89.82} & 47.21 & \textbf{74.88} & \textbf{74.44} & \textbf{57.72} \\
      \hline
      Ours ($p4$) & \textbf{89.09} & \textbf{77.97} & 62.28 & \textbf{92.39} & 97.25 & 85.22 & \textbf{0.64} & 30.30 & \textbf{49.27} & 45.61 & 77.53 & 85.69 & \textbf{47.60} & 70.15 & 73.78 & 54.26\\
      Ours ($p4m$) & \textbf{87.83} & \textbf{75.54} & 60.71 & 91.97 & 96.41 & 83.66 & \textbf{0.21} & 32.82 & 48.21 & 43.53 & 72.59 & 83.28 & \textbf{48.01} & 68.71 & 66.97 & 52.89 \\
      \hline
    \end{tabular}
  }
  \caption{Evaluation results on the S3DIS Area-5 (1-fold) in overall accuracy (OA, \%), micro-averaged accuracy (mAcc, \%), micro-averaged IoU (mIoU, \%) and per-class IoU (\%).}
  \label{tab:S3DIS-A5}
\end{table*}

\begin{table*}[h!]
  \centering
  \scalebox{0.71}{
        \begin{tabular}{  l | c | c | c | c | c | c | c | c | c | c | c | c | c | c | c | c }
             \cline{1-17}
             Method & OA & mAcc & mIoU & ceiling & floor & wall & beam & column & window & door & table & chair & sofa & bookcase & board & clutter \\
      \hline
            PointNet~\cite{Qi2017a} & 78.5 & 66.2 & 47.6 & 88.0 & 88.7 & 69.3 & 42.4 & 23.1 & 47.5 & 51.6 & 54.1 & 42.0 & 9.6 & 38.2 & 29.4 & 35.2 \\
            SPGraph~\cite{Landrieu2017} & 85.5 & 73.0 & 62.1 & 89.9 & 95.1 & 76.4 & 62.8 & 47.1 & 55.3 & 68.4 & 73.5 & 69.2 & \textbf{63.2} & 45.9 & 8.7 & 52.9 \\
            RSNet~\cite{huang2018recurrent} & - & 66.45 & 56.47 & 92.48 & 92.83 & 78.56 & 32.75 & 34.37 & 51.62 & 68.11 & 60.13 & 59.72 & 50.22 & 16.42 & 44.85 & 52.03 \\
            VV-Net~\cite{meng2019vv} & 81.45 & - & 55.07 & 83.27 & 93.95 & 59.37 & \textbf{64.35} & 40.23 & 54.06 & 66.48 & 65.20 & 63.52 & 41.48 & 20.37 & 16.21 & 47.41 \\
      PointCNN~\cite{Li2018} & 88.14 & 75.61 & 65.39 & \textbf{94.78} & \textbf{97.30} & 75.82 & 63.25 & \textbf{51.71} & \textbf{58.38} & 57.18 & \textbf{71.63} & 69.12 & 39.08 & 61.15 & 52.19 & 58.59 \\
            A-CNN~\cite{komarichev2019acnn} & 87.3 & - & 62.9 & 92.4 & 96.4 & 79.2 & 59.5 & 34.2 & 56.3 & 65.0 & 66.5 & 78.0 & 28.5 & 56.9 & 48.0 & 56.8 \\
      \hline
      Ours ($p4$) & \textbf{90.06} & \textbf{81.30} & \textbf{68.01} & 93.38 & 96.33 & \textbf{84.28} & 42.37 & 47.88 & 49.86 & \textbf{76.66} & 70.02 & \textbf{80.29} & 60.20 & \textbf{62.22} & \textbf{57.09} & \textbf{63.57} \\
      Ours ($p4m$) & \textbf{88.35} & \textbf{82.46} & \textbf{66.98} & 92.56 & 95.70 & \textbf{80.19} & 53.89 & 48.49 & 51.05 & \textbf{70.76} & 69.04 & \textbf{80.59} & 57.34 & 60.33 & 51.98 & \textbf{58.76} \\
      \hline
    \end{tabular}
  }
  \caption{Evaluation results on the S3DIS (6-fold) in overall accuracy (OA, \%), micro-averaged accuracy (mAcc, \%), micro-averaged IoU (mIoU, \%) and per-class IoU (\%).}
  \label{tab:S3DIS}
\end{table*}

\subsubsection{Segmentation} We evaluate the segmentation performance on the ShapeNet-part \cite{shapenet2015} dataset using a 14,007 / 2,874 training / testing split. We use the dataset provided by \cite{Qi2017a} and randomly sample 2,048 points with their normals as our input point cloud. For point cloud which originally has less than 2,048 points, we pad them into 2,048 points using random duplication and assume the category for each point cloud is priorly known and the irrelevant predictions are masked following the practice in \cite{Li2018,Graham2018}. 
The testing results on ShapeNet-part dataset are presented in Table \ref{tab:ShapeNet}. Compared with other works, we are on-par with the state-of-the-art performance. Interestingly, DV-ConvNet performs quite well for inputs with symmetrical shapes, such as “rocket” and “cap”, where all the other methods yield less satisfactory results.

We also evaluate the segmentation performance on S3DIS \cite{armeni_cvpr16}, a large-scale dataset for in-door scene point cloud semantic segmentation . We crop each room into $1.5\times 1.5$ $m^2$ tiles and pad an additional 0.2 m offset at the surroundings for each tile. The points in the padding area only provide additional ambient information and they are not linked with the loss calculation in both training and testing phases. We randomly sample 4,096 points (including offset) with their RGB values as our input. Our evaluation is conducted based on both 1-fold and 6-fold cross validation, and the results are shown in Table \ref{tab:S3DIS-A5} and Table \ref{tab:S3DIS}, respectively.

The $p4$ group based DV-ConvNet reaches the state-of-the-art performance regarding all the three main evaluation indices for 6-fold cross validation, and it also achieves the best IoU performance for 6 out of 13 categories. Compared with the previous state-of-the-art PointCNN \cite{Li2018}, our proposed DV-ConvNet yields much higher prediction accuracy, boosting the $mAcc$ index by 9.5\% at most. We compare to the VV-Net equipped with group convolution only, thus it has the same voxelized input as ours. However, compared to its VAE equipped version, our method still overperforms it regarding the mpIoU index in ShapeNet (84.6\% vs. 84.2\%) and the OA index in S3DIS (90.06\% vs. 87.78\%).

\subsection{Efficiency of DV-ConvNet}
\subsubsection{Computation efficiency} To better illustrate the computational efficiency of DV-ConvNet, we compare the runtime statistics between our method and other works. The statistics are summarized in Table. \ref{tab:stat}, calculated based on the same hardware configuration with a batch size of 32 during the training phase. Through the comparison, our DV-ConvNet achieves the best performance on ModelNet40 (Table. \ref{tab:ModelNet40}), while using the least FLOPs and processing time.

\begin{table}[h]
  \centering
  \scalebox{0.6}{
    \begin{tabular}{l | c | c | c | c | c | c }
      \cline{1-5}
      \hline
      Network & P-wise Conv & PointNet++ & PointCNN  & Minkowski & Ours ($p4$) & Ours ($p4m$) \\
      \hline
      \# params & 4.52M & 1.48M & \textbf{0.6M} & 2.43M & 1.62M & 1.01M \\
      \# FLOPs & 673M & 1684M & 743M & 634M & \textbf{146M} & \textbf{279M} \\
      Time & 0.94s & 0.178s & 0.06s & 0.143s & \textbf{0.054s} & \textbf{0.058s} \\
      \hline
    \end{tabular}
  }
  \caption{Number of parameters, FLOPs and running time comparisons.}
  \label{tab:stat}
\end{table}

PointNet++ \cite{Qi2017} and PointCNN \cite{Li2018} take raw point cloud coordinates as input, even though they have a relatively small number of parameters, they still involve massive FLOPs, since they employ rather complex MLPs based network architectures to learn shape representations from the point coordinates. PCNN \cite{wang2018deep} is a continuous convolution based method, as aforementioned, with the expensive polynomial based convolution kernel, it has the highest processing time cost among all the methods. For the Minkowski-ConvNet \cite{choy20194d}, we do not follow the architecture suggested (ModelNet40 evaluation is not reported in its original paper), since it has a rather complicated ResNet based architecture. Instead, we setup a similar 4-layer architecture as our DV-ConvNet for fair comparison, and based on this configuration, we notice that Minkowski-ConvNet can only achieve ~88\% OA on ModelNet40. We believe it encounters severe quantization artifact when the resolution is not sufficient. Increasing resolution will lead to a better performance, but more FLOPs will be involved as well.

\subsubsection{Converging speed} 
\begin{figure}[h!]
\includegraphics[width=0.9\columnwidth]{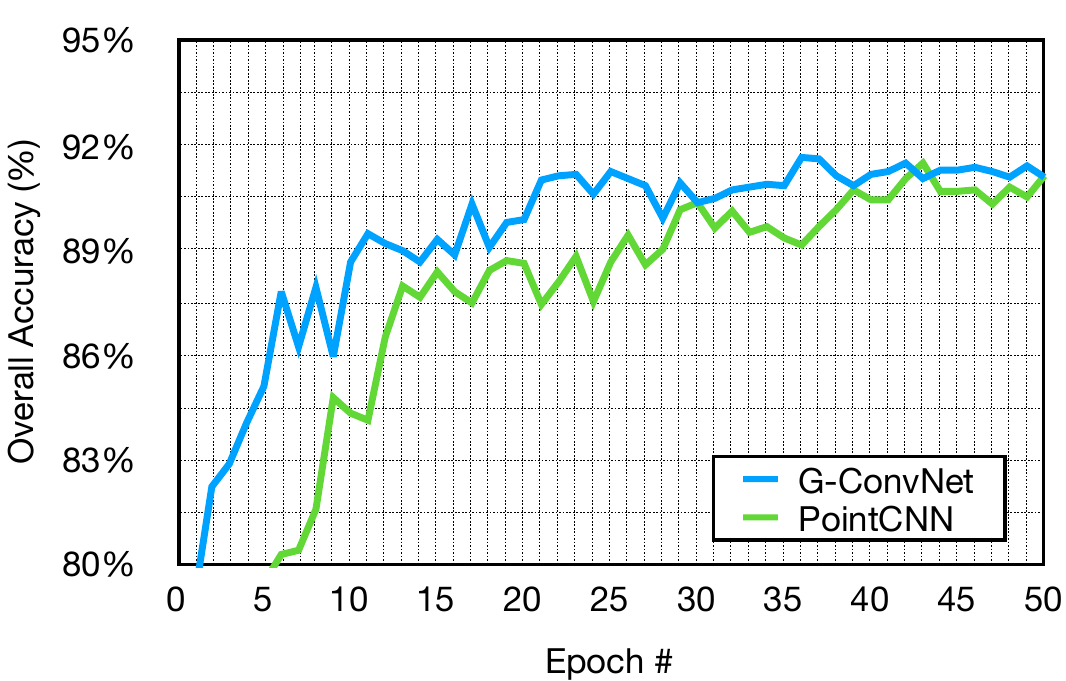}
\centering
{\caption{Epoch-accuracy curve of G-ConvNet and PointCNN, based on ModelNet40 under "Unaligned" mode.}\label{compare}}
\end{figure}

We plot the overall accuracy (OA) against training epoch curve of DV-Conv and PointCNN \cite{Li2018} in Fig. \ref{compare}, based on the ModelNet40 under the "unaligned" evaluation mode. Since the shape information are explicitly preserved by voxels, our DV-ConvNet is able to converge at a considerably faster speed, leading to more efficient model development and tuning procedure. DV-ConvNet achieves 90\% OA after about 20 training epochs, while by contrast, PointCNN takes about 40 epochs.

\subsection{Ablation Study}
All the ablation studies below are based on the ModelNet40 dataset with "Pre-aligned" evaluation mode and $p4$ group convolution enabled.

\subsubsection{Impact of $K$ and $D$.}
During dynamic voxelization operation, a grid is assigned with multiple candidate points (any points falling into that grid), and the channel-wise maximum among all the candidate points is chosen based on the max-pooling operation. By doing so, we enlarge the receptive field while keeping the computation expense within a rational range and avoid losing critical information. 

$K\times D$ controls the size of receptive field. Here we study the impact of $K$ and $D$ on the model performance by increasing one and decreasing the other accordingly, so that the relative receptive field in each convolution layer remains unchanged. Evaluation results on ModelNet40 are shown as tests 1-4 in Table \ref{tab:Ablation}. The best performance is achieved when $D=2$. 

\begin{table}[h]
    \centering
  \scalebox{0.75}{
    \begin{tabular}{ c | c | c | c | c | c | c}
            \hline
            Test \# & Group & Sampling & Pooling & $K$ & $D$ & Acc (\%) \\
      \hline
      1 & $p4$ & KNN & Max & 64. & 1 & 92.9  \\
      2 & $p4$ & KNN & Max & 32 & 2 & \textbf{93.3}  \\
      3 & $p4$ & KNN & Max & 21 & 3 & 93.0  \\
            4 & $p4$ & KNN & Max & 16 & 4 & 92.4 \\
            5 & $p4$ & KNN & Average & 32 & 2 & 92.7  \\
            6 & $p4$ & Fixed & Max & - & - & 91.9  \\
            7 & -&KNN & Max & 32 & 2 & 92.6  \\
            \hline
    \end{tabular}
  }
  \caption{Evaluation results on ModelNet40 with different model configurations.}
  \label{tab:Ablation}
\end{table}

\subsubsection{Impact of different sampling/pooling methods} Apart from the max-pooling methods, we also test the performance based on average pooling, where the channel-wise averages are used. The results are shown as test 5 in Table \ref{tab:Ablation}, and we find that the performance drops by 0.6\% when max-pooling is replaced by average-pooling. In addition, the sampling strategy using fixed sampling radius instead of k-NN is used, where the sampling radius is kept the same as the relative receptive field of each convolution layer (shown as test 6 in Table \ref{tab:Ablation}). Again, there is an obvious performance decrease, and this harmonizes well with the intuition: voxelization with self-adaptive resolution fits better for those points inhomogeneously distributed in space.

\subsubsection{Effectiveness of group convolution}
Apart from the $p4$ and $p4m$ based DV-ConvNet models, we also evaluate the performance on a baseline model to verify the effectiveness of group convolution. Our baseline model shares the same network architecture with DV-ConvNet, while the only difference is that all the group convolutions are replaced by plain 3D convolutions. The comparison between test 2 and test 7 further verifies that our DV-ConvNet is indeed able to benifit from group convolution techniques.

\subsection{Experiment Details}
Dropout with a ratio of 0.5 was used for all the MLPs in the network, except for the last layer. We chose the ADAM optimizer \cite{Kingma2014} with an initial learning rate of 0.001, and the learning rate was decreased by a factor of 0.8 for each 10 successive training epochs. $L2$ regularization was also used with a weight decay of 1e-5. 

We built up our networks based on TensorFlow \cite{Abadi2016} and the experiments in this paper were carried out based on two NVIDIA RTX 2080Ti GPUs using CUDA 10.0. We adopted a batch size of 16 for ModelNet40 classification and 32 for all the segmentation tasks.


\section{Conclusion}

In this work, we present a fully-convolutional network architecture DV-ConvNet for point cloud interpretaion. DV-ConvNet deals with the sparsity of point cloud in an efficient fashion by hierarchically converting the regional sub-portions of a point cloud into voxels. The introduction of group convolution further exploits the symmetry of the input point cloud and leads towards an efficient parameter utilization, leading to state-of-the-art performance on several benchmark datasets. DV-ConvNet solves the general and fundamental feature extraction problem in 3D point cloud neural nets, and we expect these ideas can be readily applicable for other 3D tasks, e.g., object detection, object tracking, point cloud registration and scene flow estimation etc.

\vfill

\pagebreak
\small
\bibliography{reference}


\end{document}